\documentclass[10pt,twocolumn,letterpaper]{article}

\usepackage{cvpr}
\usepackage{times}
\usepackage{epsfig}
\usepackage{graphicx}
\usepackage{amsmath}
\usepackage{amssymb}
\usepackage{array}
\usepackage{multirow, nicefrac}
\usepackage{pifont}

\usepackage[margin=4pt,font=footnotesize,labelfont=bf,tableposition=top]{caption}

\usepackage[pagebackref=true,breaklinks=true,letterpaper=true,colorlinks,bookmarks=false]{hyperref}

\cvprfinalcopy 

\def\cvprPaperID{8582} 

\usepackage{xspace}

\ifcvprfinal\fi
\begin{document}

\title{ParC-Net: Position Aware Circular Convolution with Merits from ConvNets and Transformer}

\author{
Haokui Zhang$ ^{\dag} $,
~ Wenze Hu$^\dag$,
~ Xiaoyu Wang$^\dag$
\\
$^\dag$ Intellifusion, Shenzhen, China
}

\maketitle
\thispagestyle{empty}

\begin{abstract}
Recently, vision transformers started to show impressive results which outperform large convolution based models significantly. However, in the area of small models for mobile or resource constrained devices, ConvNet still has its own advantages in both performance and model complexity. We propose ParC-Net, a pure ConvNet based backbone model that further strengthens these advantages by fusing the merits of vision transformers into ConvNets. Specifically, we propose \textbf{\emph{p}}osition \textbf{\emph{a}}ware ci\textbf{\emph{r}}cular \textbf{\emph{c}}onvolution (ParC),, a light-weight convolution op which boasts a global receptive field while producing location sensitive features as in local convolutions. We combine the ParCs and squeeze-exictation ops to form a meta-former like model block, which further has the attention mechanism like transformers. The aforementioned block can be used in plug-and-play manner to replace relevant blocks in ConvNets or transformers. Experiment results show that the proposed ParC-Net achieves better performance than popular light-weight ConvNets and vision transformer based models in common vision tasks and datasets, while having fewer parameters and faster inference speed. For classification on ImageNet-1k, ParC-Net achieves 78.6\% top-1 accuracy with about 5.0 million parameters, saving 11\% parameters and 13\% computational cost but gaining 0.2\% higher accuracy and 23\% faster inference speed (on ARM based Rockchip RK3288) compared with MobileViT, and uses only 0.5$\times$ parameters but gaining 2.7\% accuracy compared with DeIT. On MS-COCO object detection and PASCAL VOC segmentation tasks, ParC-Net also shows better performance. Source code is available at \url{https://github.com/hkzhang91/ParC-Net} 
\end{abstract}


\section{Introduction}

Recently, various vision transformers (ViTs) models have achieved remarkable results in many vision tasks, forming strong alternatives to convolutional neural networks (ConvNets)~\cite{dosovitskiy2020image}~\cite{touvron2021training}~\cite{liu2021swin}.

However, we believe both ViTs and ConvNets are indispensable for the following reasons:  1) From application perspective, both ViTs and ConvNets have their advantages and disadvantages. ViT models generally have better performance but usually suffer from high computational cost and are difficult to train~\cite{touvron2021training}. Compared with ViTs, ConvNets may show inferior performance, but they still have some unique advantages. For instance, ConvNets have better hardware support and are easy to train. In addition, as is summarized in \cite{guo2021cmt} and our experiments, ConvNets still dominate in the area of small models for mobile or edge devices. 2) From the information processing perspective, both ViTs and ConvNets have unique features. ViTs are good at extracting global information and use attention mechanism to extract information from different locations driven by input data~\cite{chen2021mobile}~\cite{mehta2022mobilevit}. ConvNets focus on modeling local relationships and have strong prior by inductive bias~\cite{dai2021coatnet}. The above analysis naturally raise a question: \emph{can we learn from ViTs to improve ConvNets for mobile or edge computing applications?}. 

In this paper, we aim to design new light-weight pure ConvNets that further enhance its strength in the area of mobile and edge computing friendly models. Pure convolution is more mobile friendly because convolutions are highly optimized by existing tool chains that are widely used to deploy model into these resource constrained devices. Even more, because of the huge popularity of ConvNets in the past few years, some existing neural network accelerators are designed mainly around convolution style operations, and the complex non-linear operations such as softmax and data bus bandwidth demanding large matrix multiplications are not efficiently supported. These hardware and software constraints make a pure convolutional light-weight model more preferable even if a ViT based model is equally competitive in other aspects.

To design such a ConvNet, we compare ConvNets with ViTs and summarize three main differences between them: 1) ViTs are good at extracting global features~\cite{chen2021mobile}~\cite{mehta2022mobilevit}~\cite{dai2021coatnet}; 2) ViTs adopt Meta-former block~\cite{yu2021metaformer}; 3) Information aggregations in ViTs are data driven (data dependent dynamic computation). Corresponding to these three points, we design our ParC block. 1) We propose the position aware circular convolution (ParC) to extract global features; 2) Based on the proposed ParC, we build a pure ConvNet Meta-former block as the basic outer structure; 3) We add channel wise attention module to the feature forward network (FFN) part of meta-former, which makes our proposed ParC block adapt kernel weights according inputs. Finally, inspired by CoatNet~\cite{dai2021coatnet} and MobileViT~\cite{mehta2022mobilevit}, we use a bifurcate structure (section~\ref{sec:ParC-Net}) as the outer frame to build a complete network ParC-Net. 

Experiment results show that the proposed ParC-Net achieves solid performance on three popular vision tasks, including image classification, object detection and semantic segmentation. Taking experiment results of image classification as an example,  ParC-Net achieves 78.6\% top-1 accuracy with about 5.0 million parameters, saving 11\% parameters and 13\% computational cost  but  gaining  0.2\%  higher  accuracy  and  23\% faster inference  speed (on Rockchip RK3288) compared with MobileViT~\cite{mehta2022mobilevit}. For experiments of object detection and semantic segmentation, compared with other light-weight models, the proposed ParC-Net achieves higher mAP and mIOU, while having fewer parameters. 

Our main contributions are summarized as follows:

\begin{itemize}
\item To overcome the restriction that traditional convolutions have limited perception fields, we propose position aware circular convolution (ParC), where base-instance kernel and position embedding strategies are used to handle input size variations and inject location information to output feature maps respectively. We jointly use the proposed ParC and conventional convolution operations to extract local-global features, which brings higher accuracy.

\item We propose ParC-Net, a pure ConvNet for mobile and edge computing applications. The proposed ParC-Net inherits advantages of ConvNets and ViTs. To our knowledge, this is the first attempt that combines strengths of ConvNets and ViTs to design a light-weight ConvNet. 

\item We apply the proposed ParC-Net on three vision tasks. Compared with the baseline model, the proposed ParC-Net achieves better performance on all three tasks, while having fewer parameters, lower computational cost and higher inference speed. 

\end{itemize}

\section{Related work}

\subsection{Vision transformers}\label{sec::vit}

Vaswani et al. firstly proposed transformer \cite{vaswani2017attention} for natural language processing (NLP) tasks. Compared with recurrent neural network (RNN) models, transformer has much higher computational efficiency and it is good at capturing relationship from any pair of elements in the input sequence. As a result, transformers replaced RNNs and dominate the NLP field. 

In 2020, Dosovitskiy et al. introduced transformer into vision tasks and proposed vision transformer (ViT)~\cite{dosovitskiy2020image}, where each image is cropped into a sequence of patches to meet the input requirement of transformer and PE is adopted to ensure the model is sensitive to position information of the input patches. With pre-training on huge datasets such as JFT-300M~\cite{sun2017revisiting}, ViT achieves impressive performance on various vision tasks. However, the original ViT model has some restrictions, for instance, it is heavy-weight, having low computational efficiency and hard to train. Subsequent variants of ViTs are proposed to overcome these problems. From the point of improving training strategy, Touvron et al.~\cite{touvron2021training} proposed to use knowledge distillation to train ViT models, and achieved competitive accuracy with less pre-training data. To further improve the model architecture, some researchers attempted to optimize ViTs by learning from ConvNets. Among them, PVT~\cite{wang2021pyramid} and CVT~\cite{wu2021cvt} insert convolutional operations into each stage of ViT model to reduce the number of tokens, and build hierarchical multi-stage structures. Swin transformer~\cite{liu2021swin} computes self attention within shifted local windows. PiT \cite{heo2021rethinking} jointly use pooling layer and depth wise convolution layer to achieve channel multiplication and spatial reduction.  CCNet~\cite{huang2019ccnet} propose a simplified version of self attention mechanism criss-cross attention and inserted it into ConvNet to build ConvNet which has global receptive field. These papers clearly show that some techniques of ConvNets can be applied on vision transformers to design better vision transformer models. 

\begin{figure*}[t]
\centering
\includegraphics[height=10cm]{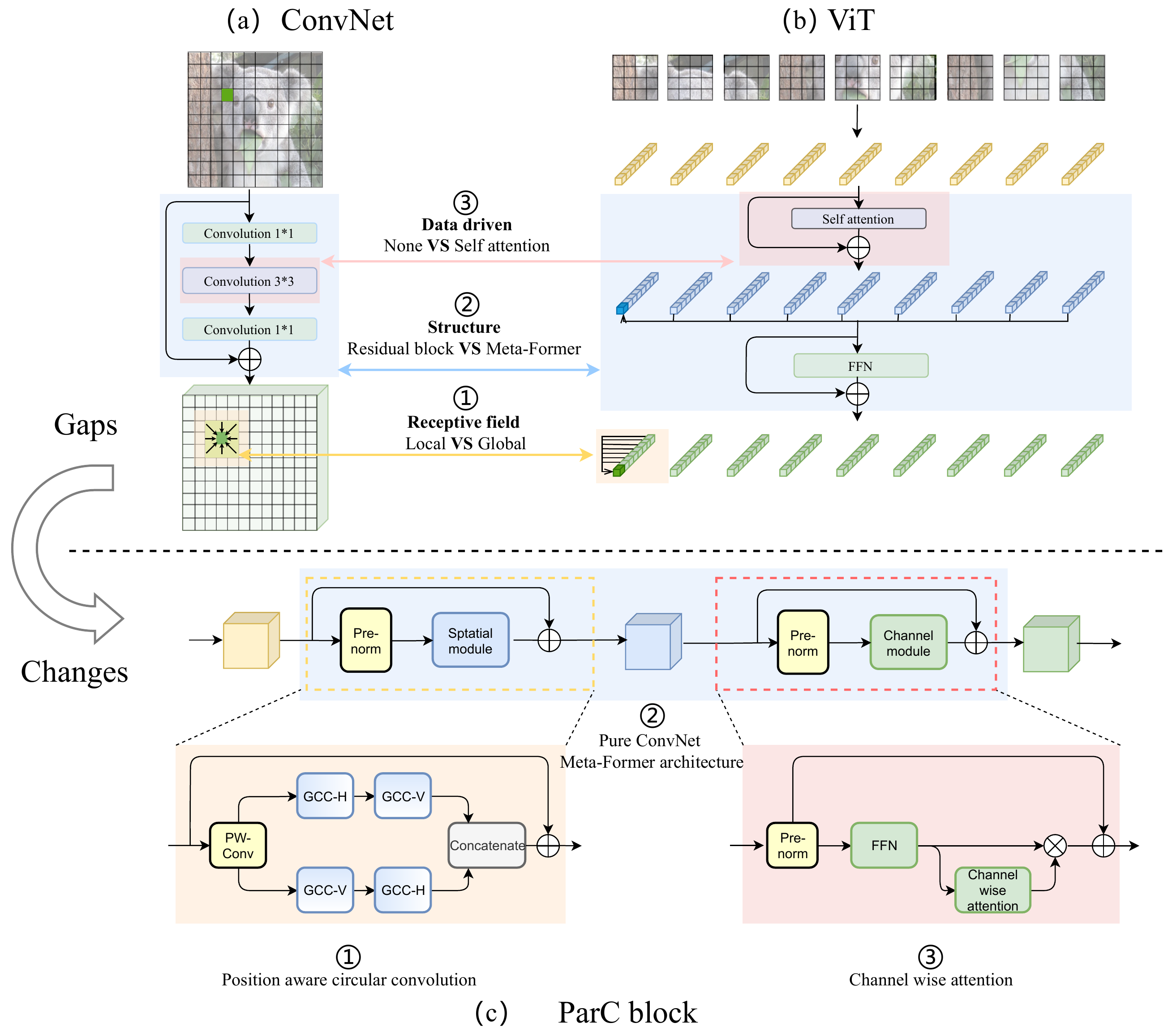}
\vspace{-0.3cm}
\caption{ParC block. (a) A residual block that is widely used in ConvNets; (b) A ViT block; (c) An ParC block}
\vspace{-0.3cm}
\label{fig:ParC_block}
\end{figure*}

\subsection{Hybrid structures combining ConvNet and vision transformers}

Another popular line of research is combining elements of ViTs and ConvNets to design new backbones. Graham et al. mixed ConvNet and transformer in their LeVit model, which significantly outperforms previous ConvNet and ViT models with respect to the speed/accuracy tradeoff~\cite{graham2021levit}. BoTNet~\cite{srinivas2021bottleneck} replaces the standard convolution with multi-head attention in the last several blocks of ResNet. ViT-C~\cite{xiao2021early} adds early convolutional stem to vanilla ViT. ConViT~\cite{d2021convit} incorporates soft convolutional inductive biases via a gated positional self-attention. The CMT~\cite{guo2021cmt} block consists of depth wise convolution based local perception unit and a light-weight transformer module. CoatNet~\cite{dai2021coatnet} merges convolution and self-attention to design a new transformer module, which focuses on both local and global information. After comprehensive comparison, we find that these hybrid models simultaneously employed similar structure, that is using convolutional stem to extract local features in the beginning stages and transformer style models later to extract global or local-global features. We choose a similar structure when designing our pure convolutional model.

\subsection{Light-weight ConvNets and ViTs}
\label{sec:light-weight}

Since 2017, light-weight ConvNets attract much attentions as more and more applications needs to run ConvNet models on mobile devices. Now, there are a lot of light-weight ConvNets, such as ShuffleNets~\cite{ma2018shufflenet}~\cite{ma2018shufflenet}, MobileNets~\cite{howard2017mobilenets}~\cite{sandler2018mobilenetv2}~\cite{howard2019searching}, MicroNet~\cite{li2021micronet}, GhostNet~\cite{han2020ghostnet}, EfficientNet~\cite{tan2019efficientnet}, TinyNet~\cite{chen2019tinynet} and MnasNet~\cite{tan2019mnasnet}. Compared with standard ConvNets, light-weight ConvNets have fewer parameters, lower computational cost and faster inference speed. In addition, light-weight ConvNets can  be applied on a wide range of devices. Despite these benefits, these light-weight models have inferior performance compared with heavy-weight models. Very recently, following the research line of combining strengths of ConvNet and ViT, some researcher attempted to build light-weight hybrid models for mobile vision tasks. Mobile-Former presents a parallel design of MobileNet and transformer, which leverages the advantages of MobileNet at extracting local features and transformer at capturing global information~\cite{chen2021mobile}. Mehta and Rastegari proposed MobileViT, where the upper stages of MobileNetv2~\cite{sandler2018mobilenetv2} are replaced with MobileViT block~\cite{mehta2022mobilevit}. In MobileViT block, local representations extracted by convolution and global representations are concatenated to generate local-global representations. 

In terms of purpose, our proposed ParC-Net is related to Mobile-Former and MobileViT. Different from these two models which still keep transformer blocks, our proposed ParC-Net is pure ConvNet, which makes our proposed ParC-Net more mobile friendly. Our experiments of deploying models on low power platform confirm this point. In terms of designing a pure ConvNet via learning from ViTs, our work is most closely related to a parallel work ConvNext~\cite{liu2022convnet}. The two major differences are: 1) Ideas and architectures are different. The ConvNext modernizes a standard ResNet toward the design of a vision transformer by introducing a series (more than ten) of incremented but effective designs. Our proposed ParC-Net starts from three main differences between ConvNets and ViTs and fills the gaps from macro level. As the ideas are different, the corresponding structures are also different; 2) They are proposed for different purposes. Our ParC-Net is proposed for mobile devices. Compared with ConvNext, the proposed ParC-Net shows advantages when constraining models as light-weight models. 

\begin{figure*}[t]
\centering
\includegraphics[height=3.8 cm]{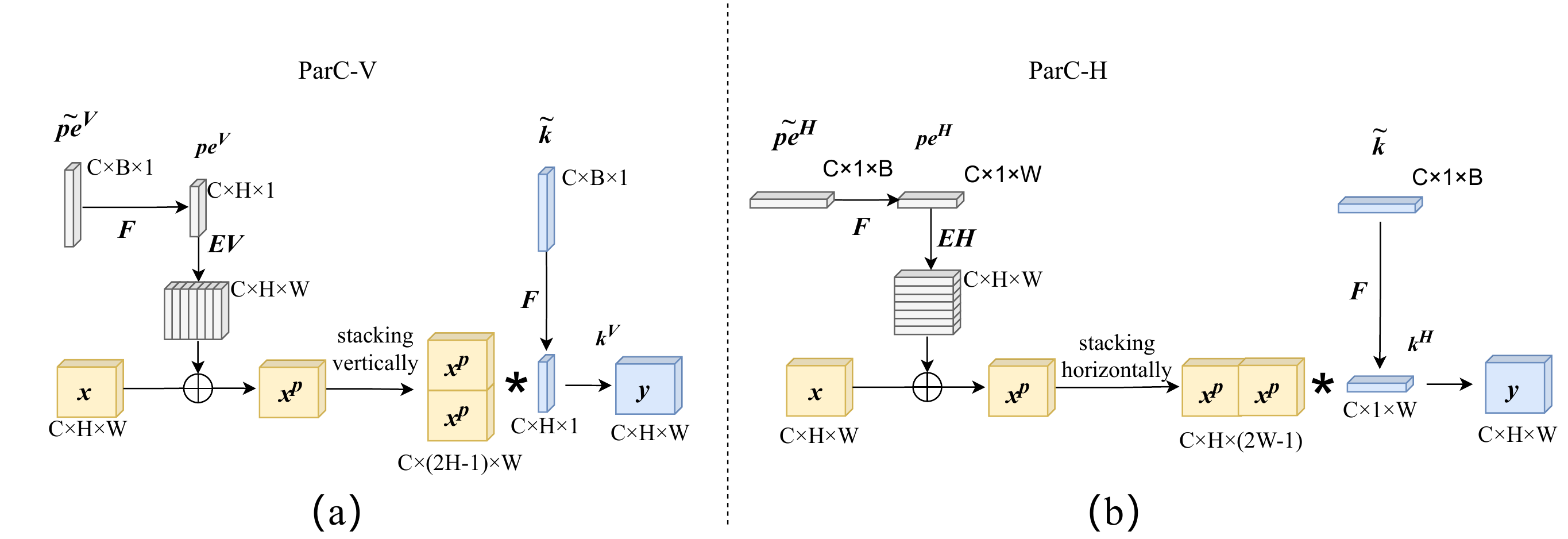}
\vspace{-0.2cm}
\caption{Illustration of the position aware circular convolution. (a) ParC-V; (b) ParC-H. $F$, $EV$ and $EH$ are explained in equations~\ref{equ:ParC_v} and~\ref{equ:ParC_h}}
\vspace{-0.2cm}
\label{fig:ParC}
\end{figure*}

\begin{figure*}[t]
\centering
\includegraphics[width=9cm]{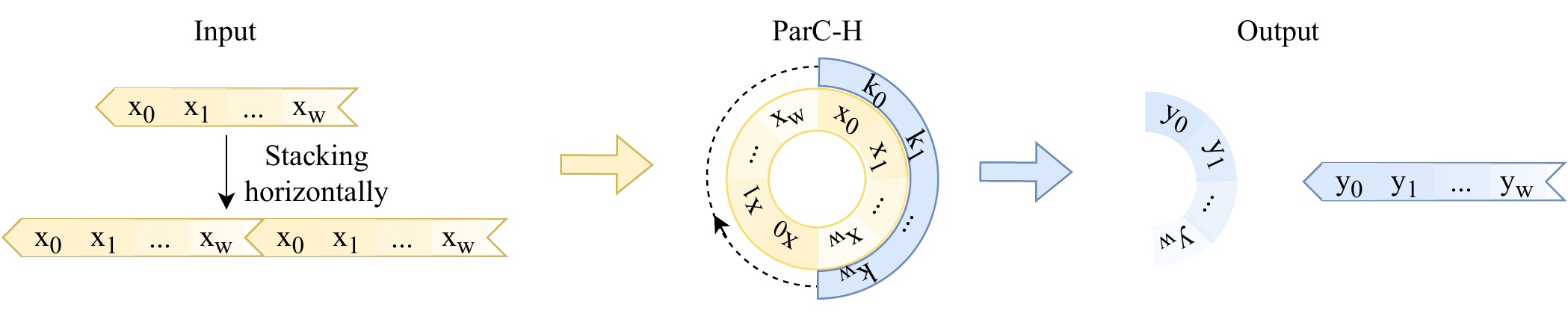}
\vspace{-0.2cm}
\caption{Illustration of global circular convolution on horizontal direction.}
\vspace{-0.2cm}
\label{fig:circular_conv}
\end{figure*}

\section{The proposed method}

In this section, we will introduce our ParC-Net in two parts, the details of the building block (ParC block) and the overall model structure (ParC-Net). 

\begin{figure*}[t]
\centering
\includegraphics[width=16 cm]{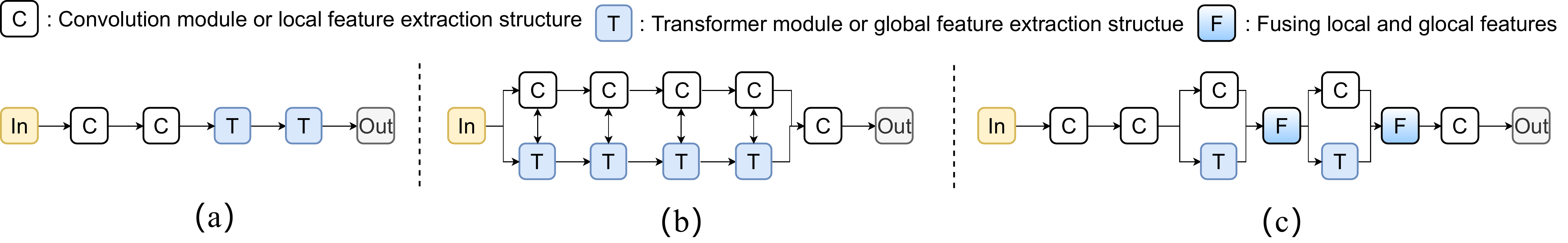}
\vspace{-0.3cm}
\caption{Three main hybrid structures. (a) serial structure; (b) parallel structure; (c) bifurcate structure}
\vspace{-0.3cm}
\label{fig:framework}
\end{figure*}

\subsection{ParC block}
\label{sec: ParC_block}

The upper part of Figure~\ref{fig:ParC_block} shows three major differences between common ConvNets and ViTs. The bottom half of Figure~\ref{fig:ParC_block} illustrates the architecture of our proposed ParC block. In the following, we will explain the motivation and the specific structure of each component of the proposed ParC block.

\noindent {\bf Extracting global features with ParC}. In ConvNets, feature is calculated as $y_{i} = \sum_{j\in \mathcal{L}(i)}{w_{i-j}x_{j}}$, where $x_{i}$, $y_{i}$ are the input and output at position $i$ respectively, and $\mathcal{L}(i)$ denotes a local neighborhood of $i$. In ViTs, self-attention modules extracts features based on formula $y_{i}=\sum_{j\in \mathcal{G}}{\frac{e^{(x_{i}^{T}x_{j})}}{\sum_{k\in\mathcal{G}}{e^{(x_{i}^{T}x_{k})}}}x_{j}}$, where $\mathcal{G}$ means the global spatial space. Comparing these two formulas, we can see that self attention learns global features from the entire spatial locations but convolution gathers information from a local receptive field.  

To overcome this issue, we propose the position aware circular convolution (ParC). As shown in Figure~\ref{fig:ParC}, our proposed ParC has two types, one is ParC of vertical direction (ParC-V) and the other one is ParC of horizontal direction (ParC-H). The receptive field of the ParC-V and ParC-H covers all pixel in the same column and the same row, respectively. Jointly using ParC-V and ParC-H can extract global features from all input pixels. For notational simplicity, we assume the input $x$ has only one channel and the corresponding shape is $1 \times h\times w$. The output of ParC-V at location $(i,j)$ is computed with:
\begin{equation}
\begin{split}
{}&    pe^{V} = F(\widetilde{pe}^{V}) = [pe_{0}^{V}, pe_{1}^{V}, \cdots, pe_{ h-1}^{V}]^{T} \\
{}&    pe_{e}^{V} = EV(pe^{V}, w)    \\
{}&    k^{V} = F(\widetilde{k}) = [k_{0}^{V}, k_{1}^{V}, \cdots, k_{h-1}^{V}]  \\
{}&    x^{p} = x + pe_{e}^{V}     \\
{}&    y_{i,j} = \sum_{t\in(0,h-1)}{k_{t}^{V}x_{((i+t) {\rm mod} h, j )}^{p}}
\end{split}
\label{equ:ParC_v}
\end{equation}
where, $pe^{V}$ is instance position embedding (PE) and it is generated from a base embedding $\widetilde{pe}^{V}$ via bilinear interpolation function $F()$. Here $F()$ is used to adapt the size of position embedding to the size of input features. $pe_{e}^{V}$ is expanded PE. $k^{V}$ is instance kernel. $EV()$ is an expand function of vertical direction. After copying the input vector $w$ times, $EV()$ concatenates these copied vectors along horizontal direction to generate a $h\times w$-sized PE matrix. Similarly, the output of ParC-H at location $(i,j)$ can be expressed as:
\begin{equation}
\begin{split}
{}&    x^{p} = x + pe_{e}^{H}     \\
{}&    y_{i,j} = \sum_{t\in(0,w-1)}{k_{t}^{H}x_{(i, (j+t) {\rm mod} w )}^{p}}
\end{split}
\label{equ:ParC_h}
\end{equation} 
where $pe_{e}^{H} = EH(pe^{H}, h)$ and $EH()$ is an expand function. $EH()$ expands input vector along the vertical direction. Implementing the ParC in modern deep learning libraries is straightforward. Taking the most complicated part $y_{i,j} = \sum_{t\in(0,w-1)}{k_{t}^{H}x_{(i, (j+t) {\rm mod} w )}^{p}}$ as an example, it can be implemented with one line of code:
$y = F.conv2D(torch.cat(x^{p}, x^{p}, dim=3), k^{H})$.  Figure~\ref{fig:circular_conv} illustrates the computational process in the case that the input is an one dimensional vector. From Figure~\ref{fig:circular_conv}, we can see that ParC-H perform convolutions along a circle generated by connecting the start and the end of the input. So, we name the proposed convolution as the circular convolution. The proposed ParC introduces three modifications:

\begin{itemize}
\item The receptive field is increased to global spatial space. Note that, increasing the kernel size of tradition local convolution to full input size does not extract global features. In local convolution, zero padding is usually used to keep the size of convolutional feature the same with that of the input. Even if we increase the kernel size to global size, the global kernel only covers part pixels coming from input. Especially for extracting feature in edge portion, only about half of pixels that covered by global kernel are from input actual input, while others are simply zeros. 

\item The PE is used to keep the output feature sensitive to spatial location. Circular convolution can extract global features, but it disturbed the spatial structure of the original input. For classification, keeping spatial structure may not be a big issue. But, as is shown in ablation study, for location sensitive tasks such as segmentation and detection, keeping spatial structure does matter. Here, following the design in ViTs, we introduce PE to keep spatial structure. Experiment results in ablation study show that PE is useful in segmentation and detection tasks

\item The kernel and PE are dynamically generated according to the input size. In ParC, the sizes of kernels and PE codes must be consistent with that of instance inputs. To handle the case that inputs have different spatial resolution, we generate instance kernels and PE codes via interpolation functions. 

\end{itemize}

\noindent {\bf Designing ParC block with ParC}.  From ConvNets to ViTs, a considerable modification is meta-former block replaced residual block (the blue two-way arrow). A Meta-former block generally consists of a sequence of two components: a token mixer and a channel mixer. The token mixer is for exchanging information among tokens in different spatial locations. The channel mixer is for mixing information among different channels. Both two components use residual learning structure. 

Inspired by this, we insert ParC into Meta-former like block to build our ParC block. Specifically, we replace self-attention module with the proposed ParC to build an new spatial module to replace token mixer part. Here, we do this for two main reasons: 1) ParC can extract global features and interacts information among pixels from global space, which meets the requirement of token mixer module; 2) the computation complexity of self attention module is quadratic. Replacing this part with ParC can reduce computational cost significantly, which is helps achieving our goal of designing a light-weight ConvNet. Based on the proposed ParC, we build a pure ConvNet meta-former like block.

\noindent {\bf Adding channel wise attention in channel mixer part}.In ViTs, self attention module can adapt weights according input, which makes ViTs data driven models. By adopting attention mechanism, data driven models can focus on important features and suppress unnecessary ones, which brings better performance. Previous literature~\cite{hu2018squeeze}\cite{woo2018cbam}\cite{jaderberg2015spatial} already explained the importance of keep model data driven. 

By replacing the self-attention with the proposed global circular convolution, we get a pure ConvNet which can extract global features. But the replaced model is no longer a data driven model. To compensate, we insert channel wise attention module into channel mixer part, as shown in Figure~\ref{fig:ParC_block}(c). Following SENet~\cite{hu2018squeeze}, we first aggregate spatial information of input features $x\in \mathbb{R}^{c\times h\times w}$ via global average pooling and get aggregated feature $x_a\in \mathbb{R}^{c\times 1\times 1}$; Then we feed $x_a$ into a multi-layer perceptron to generate channel wise weight $a\in \mathbb{R}^{c\times 1\times 1}$. The $a$ is multiplied with $x$ channel wise to generate the final output. 

\subsection{ParC-Net}
\label{sec:ParC-Net}

\begin{figure*}[t]
\centering
\includegraphics[width=16 cm]{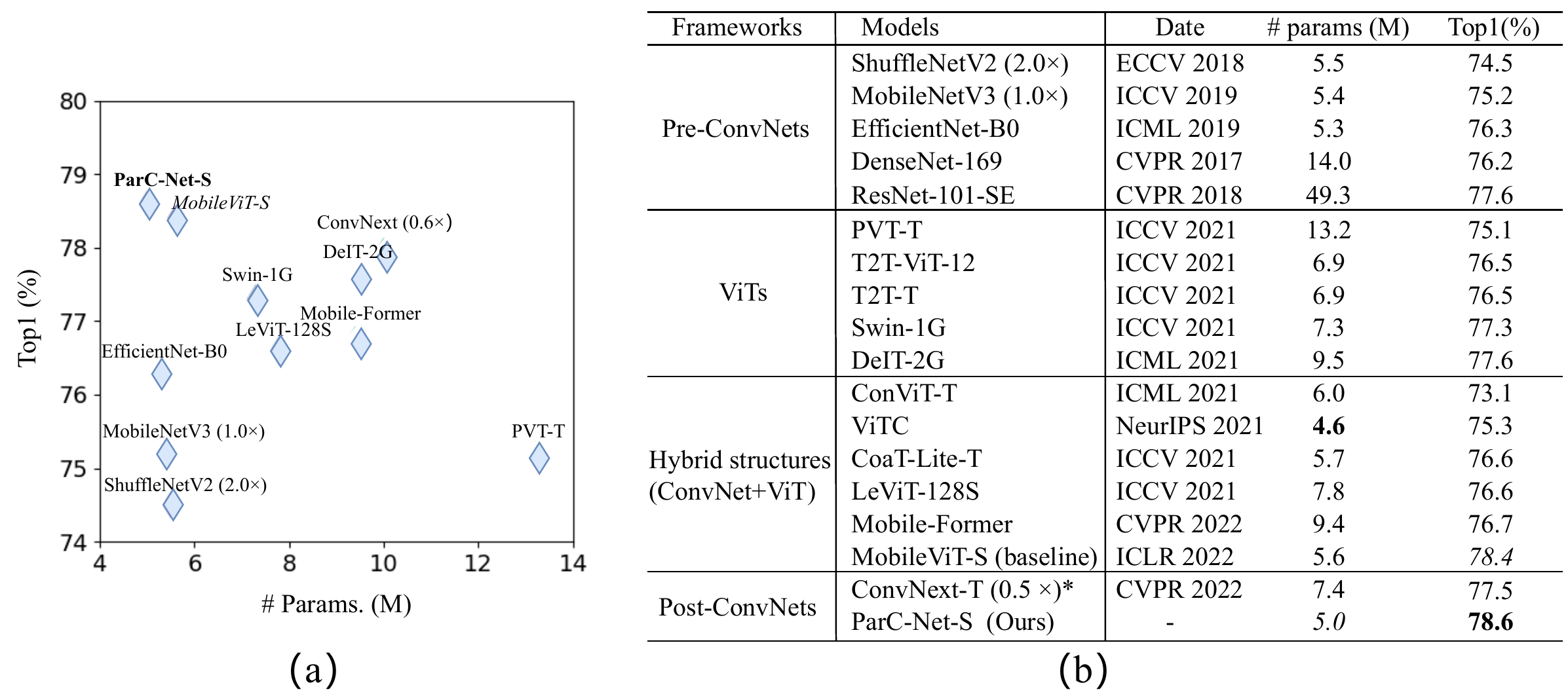}
\vspace{-0.3cm}
\caption{Classification experiment results on ImageNet-1K.  (a) Accuracy vs model size. Here we only keep part of comparison models for clarity. (b) Comparison of results on image classification. $*$ indicates our implementation. Pre-ConvNets indicate classical ConvNets appeared before ViTs. Post-ConvNets denote ConvNets which integrate merits of ViTs but still keep pure ConvNet structures.}

\label{fig:res_cls}
\end{figure*}

\begin{table*}[t]
\renewcommand\arraystretch{1.0}
\begin{center}
\begin{tabular}{l|l|l|c|c}
\hline
Frameworks  &   Models     &   Date  &   \# params (M)  &   Top1(\%)  \\
\hline
  &  ShuffleNetV2(2.0$\times$) &  ECCV 2018  &  5.5  & 74.5 \\
Pre-CNNs             &  MobileNetV3(1.0$\times$)  &  ICCV 2019  &  5.4  & 75.2 \\
                     &  EfficientNet-B0           &  ICML 2019  &  5.3  & 76.3 \\
\hline
ViTs                 &  T2T-ViT-7                 &  ICCV 2021  &  4.3  & 71.7 \\
                     &  DeiT-T                    &  ICML 2021  &  5.7  & 72.2 \\
\hline
                     &  ViT-C                      &  NeurIPS 2021 & 4.6 & 75.3 \\
Hybrid structures    &  CoaT-Lite-T               &  ICCV 2021  &  5.7  & 76.6 \\
                     &  MobileViT-S               &  ICLR 2022  &  5.6  & 78.4 \\
\hline
Post-CNN             &  ParC-Net-S              &  ~~~~~~~-          &  5.0  & 78.6 \\
\hline
\end{tabular}

\label{tab:lw_models}
\end{center}
\caption{Comparisons of light-weight models on ImageNet-1K classification}
\vspace{-0.3cm}
\end{table*}

In section~\ref{sec: ParC_block}, we have presented the ParC block, which is a basic block and can be inserted into most of the current existing models. In this section, we select an outer frame for it and build the complete network ParC-Net. 

Currently, as shown in Figure~\ref{fig:framework}, existing hybrid structures can be basically divided into three main structures, including serial structure (Figure~\ref{fig:framework}(a))~\cite{graham2021levit}\cite{xiao2021early}, parallel structure (Figure~\ref{fig:framework}(b))~\cite{chen2021mobile} and bifurcate structure (Figure~\ref{fig:framework}(c))~\cite{mehta2022mobilevit}\cite{dai2021coatnet}. Among all three structures, the third one achieves best performance for now. At present, bifurcate model CoatNet~\cite{dai2021coatnet} achieves the highest classification accuracy on Imagenet-1k. Mobile device aimed model MobileViT~\cite{mehta2022mobilevit} also adopts the third structure. 

Inspired by this, we adopt bifurcate structure as our outer frame and build our final outer frame based on MobileViT. Specifically, taking the outer frame adopted in MobileViT as baseline, we further make some improvements: 

\begin{itemize}
\item MobileViT consists of two major types of modules. Shallow stages consist of MobileNetV2 blocks, which have local receptive field. Deep stages are made up of ViT blocks, which enjoy global receptive field. We keep all MobileNetV2 blocks and replacing ViT blocks with corresponding ParC blocks. This replacement converts the model from hybrid structure to pure ConvNet. 

\item We appropriately increase the widths of ParC blocks. Even so, the replaced model still has fewer parameters and less computational cost. 
\item As show in Figure~\ref{fig:framework}(c), the bifurcate structure contains some interaction modules, which are in charge of interacting information between local and global feature modules. In the original MobileViT, ViT blocks are the most heavy modules. After replacing ViT blocks with ParC blocks, the cost of these interaction modules becomes prominent. So, we introduce group convolution and point wise convolution into these modules, which decreases number of parameters without hurting performance.

\end{itemize}

\section{Experiment results}
In experiments, we show the overall advantages of the proposed ParC-Net on three typical vision tasks, and then conduct detailed study to show the value of our design choices, the model scaling characteristics, and its speed advantage on low power devices.

\begin{figure*}[t]
\centering
\includegraphics[width=16 cm]{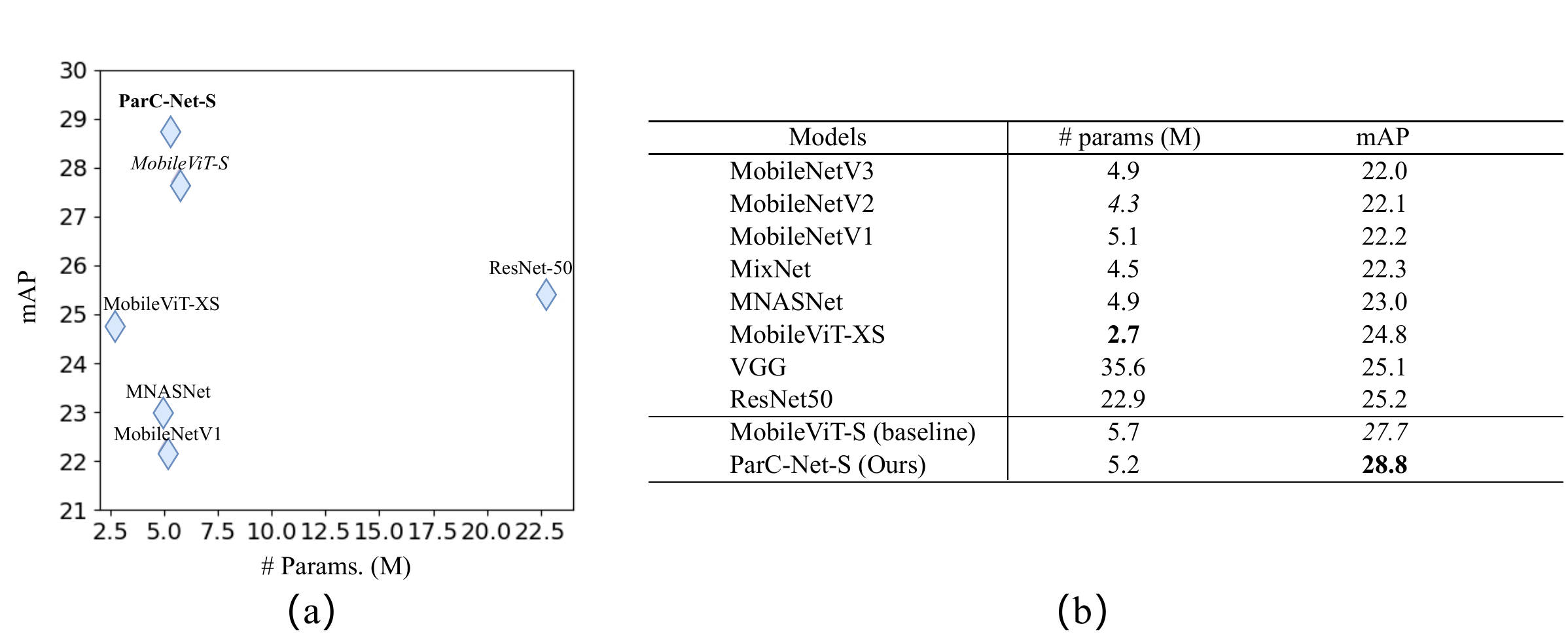}
\caption{Object detection results on MS-COCO. (a) mAP vs model size. (b) Comparison results}
\label{fig:res_det}
\end{figure*}

\subsection{Image classification}
\label{sec: img_cls}

We conduct image classification experiments on ImageNet-1k, the most widely used benchmark dataset for this task. We train the proposed ParC-Net models on the training set of ImageNet-1K, and report top-1 accuracy on the validation set.  

{\bf Training setting}. As we adopt MobileViT like structure as our outer framework, we train our model using a very similar training strategy as well. To be specific, we train each model for 300 epochs on 8 V100 or A100 GPUs with AdamW optimizer~\cite{loshchilov2019decoupled}, where the maximum learning rate, minimum learning rate, weight decay and batchsize are set to 0.004, 0.0004, 0.025 and 1024 respectively. Optimizer momentum $\beta_{1}$ and $\beta_{2}$ of the AdamW optimizer are set to 0.9 and 0.999 respectively. We use the first 3000 iterations as warm up stage. We adjust learning rate following the cosine schedule. For data augmentation, we use random cropping, horizontal flipping and multi-scale sampler. We use label smoothing~\cite{szegedy2016rethinking} to regularize the networks and set smoothing factor to 0.1. We use Exponential Moving Average (EMA)~\cite{polyak1992acceleration}. More details of the training settings and \emph{ link to source code will be provided in supplementary materials}. 

{\bf Comparison results}. The experiment results of image classification are listed in Figure~\ref{fig:res_cls}. Figure~\ref{fig:res_cls} (a) shows that ParC-Net-S and MobileViT-S beat other model by a clear margin. Figure~\ref{fig:res_cls} (b) shows comparison with more models. The proposed ParC-Net-S achieves highest classification accuracy, and have fewer parameters than most models. Compared with the second best model MobileViT-S, our ParC-Net-S decreases the number of parameters by 11\% and increases the top 1 accuracy by 0.2 percentage points.

{\bf Light-weight models}. Table~\ref{tab:lw_models} shows comparison results among light-weight models, which confirms our ideas and answers the question proposed in introduction. 

Firstly, comparing results of light-weight ConvNets with that of ViTs, light-weight ConvNets show much better performance. 

Secondly, comparing the popular ConvNets before ViT appears (pre-ConvNets), ViTs and hybrid structures, hybrid structures achieve the best performance. Therefore improving ConvNets by learning from the merits of ViT is feasible. 

Finally, the proposed ParC-Net achieves the best performance among all comparison models. So indeed by learning from ViT design, performance of pure light-weight ConvNets can be improved significantly. 

\subsection{Object detection}

We use MS-COCO~\cite{2014Microsoft} datasets and its evaluation protocol for object detection experiments. Following~\cite{mehta2022mobilevit}\cite{sandler2018mobilenetv2}, we take single shot object detection (SSD)~\cite{2016SSD} as the detection framework and use separable convolution to replace the standard convolutions in the detection head. 

{\bf Experiment setting}. Taking models pretrained on ImageNet-1K as backbone, we finetune detection models on training set of MS-COCO with AdamW optimizer for 200 epochs. Batchsize and weight decay are set to 128 and 0.01. We use the first 500 iterations as warm up stage, where the learning rate is increased from 0.000001 to 0.0009. Both label smoothing and EMA are used during training.  

{\bf Comparison results}. Figure~\ref{fig:res_det} lists the corresponding results. Similar to results in image classifiction, MobileViT-S and ParC-Net-S achieve the the second best and the best in terms of mAP. Compared with the second best model, ParC-Net-S shows advantages in both model size and detection accuracy.  

\begin{figure*}[t]
\setlength{\abovecaptionskip}{0.cm}
\setlength{\belowcaptionskip}{0.0cm}
\centering
\includegraphics[width=16 cm]{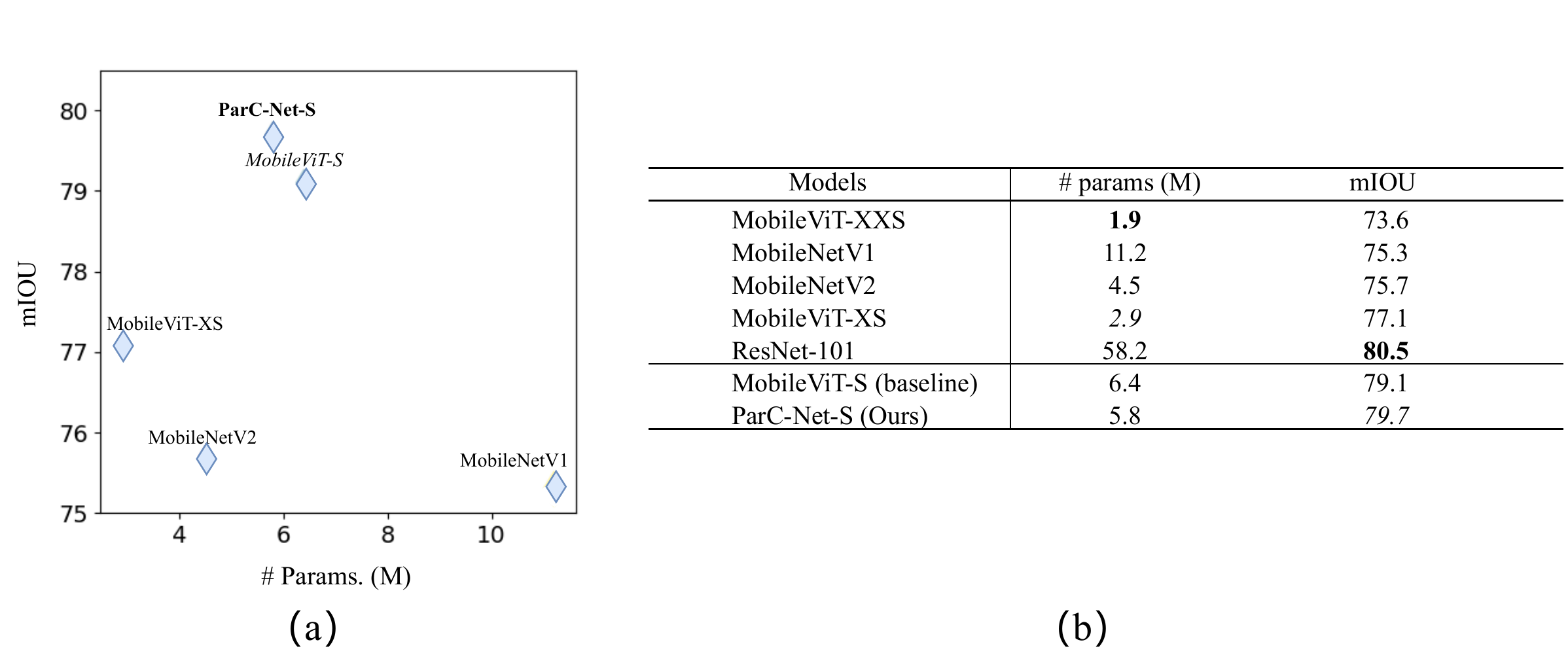}
\caption{Semantic segmentation experiments on PASCAL VOC. (a) mIOU vs model size.(b) Comparison results with more models.}
\label{fig:res_seg}
\end{figure*}

\begin{table*}
\renewcommand\arraystretch{0.9}
\begin{center}

\begin{tabular}{l|l|l|ccc|c|c}
\hline
Row & Task &   Kernel  &  MF  &  CA  & PE & \# params (M) &Top1/mAP/mIOU
\\
\hline
1 & classification &  Baseline         &   -   &   -    & -   & 5.6   & 78.35 \\
2 & classification &  BK 1/4           &   Y   &   Y    & N   & 5.0 & 78.46 \\
3 & classification &  BK 1/2           &   Y   &   Y    & N   & 5.0 & 78.45 \\
4 & classification &  ParC              &   N   &   Y    & Y   & 5.3 & 76.00 \\
5 & classification &  ParC              &   Y   &   N    & Y   & 5.0 & 78.50 \\
6 & classification &  ParC              &   Y   &   Y    & N   & 5.0 & 78.63 \\
7 & classification &  ParC              &   Y   &   Y    & Y   & 5.0 & 78.63 \\
\hline
8 & detection      &  Baseline         &   -   &   -    & -   & 5.7   & 27.7   \\
9 & detection      &  ParC              &   Y   &   Y    & N   & 5.2 & 27.5      \\
10 & detection     &  ParC              &   Y   &   Y    & Y   & 5.2   & 28.5   \\
\hline
11 & segmentation  &  Baseline         &   -   &   -    & -   & 6.4   & 79.1   \\
12 & segmentation  &  ParC              &   Y   &   Y    & N   & 5.8   & 79.2   \\
13 & segmentation  &  ParC              &   Y   &   Y    & Y   & 5.8   & 79.7   \\
  
\hline
\end{tabular}

\label{tab:ablation}
\end{center}
\caption{Ablation study. BK, MF, CA and PE denote big kernel, meta-former architecture, channel wise attention and position embedding. BK 1/4 and BK 1/2 means the kernel size is set to 1/4 and 1/2 of the input features, respectively.}
\end{table*}

\subsection{Semantic segmentation}

{\bf Experiment settings}. DeepLabV3 is adopted as the semantic segmentation framework. We fine tune segmentation models on training set of PASCAL VOC~\cite{2015The} and COCO dataset, then evaluate trained models on validation set of PASCAL VOC using mean intersection over union (mIOU) and report the final results for comparison. We fine tune each model for 50 epochs with AdamW. Readers may refer to more details about training settings in supplementary materials.

{\bf Comparison results}. Results are summarized in Figure~\ref{fig:res_seg}. We can see that MobileViT-S and ParC-Net-S have the best trade-off between model scale and mIOU. Compared with ResNet-101,  MobileViT-S and ParC-Net-S achieve competitive mIOU, while having much fewer parameters. 

\begin{table*}[t]
\renewcommand\arraystretch{1.1}
\begin{center}
{
\begin{tabular}{l|l|cccccl}
\hline
Row & Models  &  \# param & FLOPs & Devices & Speed (ms) & Top1 (\%)\\
\hline
1 & MobileViT-S       & 5.6M  & 4.0G & RK3288 & 457  & 78.4 \\
2 & ParC-Net-S        & 5.0M  & 3.5G & RK3288 & 353  & 78.6 \\
\hline
3 & MobileViT-S       & 5.6M  & 4.0G & DP2000 & 368 & 78.4 \\
4 & ParC-Net-S        & 5.0M  & 3.5G & DP2000 & 98  & 78.6 \\
\hline
5 & ResNet50~*        & 26 M  & 4.1G &  CPU   & 98   & 78.8  \\
6 & ParC-ResNet50~*   & 24 M  & 4.0G &  CPU   & 98   & 79.6  \\
\hline
7 & MobileNetV2*      & 3.5M  & 0.6G & CPU    & 24 & 70.2  \\
8 & ParC-MobileNetV2* & 3.5M  & 0.6G & CPU    & 27 & 71.1  \\
\hline
9 & ConvNext-T(0.5$\times$W)*        & 7.4M  & 1.1G & CPU  & 47 & 77.5  \\
10 & ParC-ConvNext-T(0.5$\times$W)*  & 7.4M  & 1.1G & CPU  & 48 & 78.3  \\
\hline
\end{tabular}
}
\end{center}
\caption{Applying ParC-Net designs on different backbones and comparing inference speeds of different models. CPU used here is Xeon E5-2680 v4. DP2000 is the code name of a in house unpublished low power neural network processor that highly optimizes the convolutions. *denotes the models are trained under convnext hyperparameters settings, which may not be the optimal. W means network width. Latency is measured with batch size 1. }
\label{Table: inference speed}
\end{table*}

\subsection{Ablation study}

Using the MobileViT as a baseline model, we further conduct ablation analysis on three components proposed in our ParC-Net. 

\begin{itemize}

\item {\bf Position aware circular convolution}. The proposed ParC has two major characteristics: 1) Circular convolution brings global receptive field; 2) PE keeps spatial structure information. Experiment results confirm that both characteristics are important. 1) Results in rows 1-3 show that, using big kernel can also improve accuracy, but the benefit of it reaches a saturation point when kernel size reaches a certain level. This results are consistent with the statement claimed in~\cite{liu2022convnet}. Using ParC can further improve accuracy, as shown in rows 2-3 and 6-7. 2) Introducing PE to ParC is necessary. As we explained in section~\ref{sec: ParC_block}, using circular convolution alone can indeed capture global features but it disturbs the original spatial structures. For classification task, PE has no impact (rows 6 and 7). However, for detection and segmentation tasks which are sensitive to spatial location, abandoning PE hurts performances (rows 9-10 and 12-13). 

\item {\bf Meta-former architecture}. In experiments of abandoning Meta-former architecture, we integrate ParC with the ResNeXt block~\cite{xie2017aggregated} to replace Meta-former architecture. By comparing row 4 and 7, we can see that using the proposed pure ConvNet meta-former architecture is useful. 

\item {\bf Channel wise attention}. Results in rows 5 and 7 show that using channel wise attention can improves performance. Compared with ParC, channel wise attention brings less benefit.

\end{itemize}

In summary, all three components are useful. Connecting them as a whole achieves the best performance.

\subsection{Inference speed on low power devices.}
\label{sec: inference}

In this section, we conduct experiments to verify two points: 1) as we mentioned in introduction, the ParC-Net is proposed for edge computing devices. To verify whether the proposed ParC-Net meets our requirements, we deploy the proposed ParC-Net on a widely used low power chip Rockchip RK3288 and an in house low power neural network processor DP2000, compare it with baseline.  We use ONNX~\cite{bai2019} and MNN\cite{jiang2020mnn} to port these models to chips and time each model for 100 iterations to measure the average inference speed; 2) The proposed ParC block is an plug-and-play block, it can be inserted into other models. We replaced convolutions in the last few blocks of typical CNNs with our proposed ParC (with PE and kernel generation etc.) Comparison results are listed in Table~\ref{Table: inference speed}. 

As shown in rows 1-4 of Table ~\ref{Table: inference speed}, compared with baseline, ParC-Net is 23\% faster on Rockchip RK3288 and 3.77 $\times$ faster On DP2000. Besides less FLOPs operations, we believe this speed improvement is also brought by two factors: 1) Convolutions are highly optimized by existing tool chains that are widely used to deploy models into these resource constrained devices; 2) Compared with convolutions, transformers are more data bandwith demanding as computing the attention map involves two large matrices $K$ and $Q$, whereas in convolutions the kernel is a rather small matrix compared with the input feature map. In case the bandwith requirement exceeds that of the chip design, the CPU will be left idle waiting for data, resulting in lower CPU utilization and overall slower inference speed; 

Results in rows 3-10 show that our proposed ParC-Net universally improves performances of typical light weight models. MobileViT-S has much higher FLOPs but achieves good trade-off between model size and accuracy, which excels in its own application purpose. By applying our ParC-Net designs on MobleViT-S, ParC-Net-S achieve better balance between model size, FLOPs and accuracy. Results on ResNet50, MobileNetV2 and ConvNext-T shows that models which focus on optimizing FLOPs-accuracy trade-offs can also benefit from our ParC-Net designs.

\section{Conclusion}

In this paper, for edge computing devices, we present ParC-Net, a pure ConvNet, which inherits advantages of ConvNet and integrated structure characteristics of ViT. To evaluate the performances, we apply the proposed model on three popular vision tasks, image classification, object detection and semantic segmentation. The proposed model achieves better performance on all three tasks, while having fewer parameters compared with other ConvNet, ViT and hybrid models. Experimental results on low power devices Rockchip RK3288 and our in house processor DP2000 show that the proposed ParC-Net does inherit ConvNets and it is well supported by edge computing devices.  

{\small
\bibliographystyle{ieee_fullname}
\bibliography{egbib}
}

\end{document}